# JuncNet: A Deep Neural Network for Road Junction Disambiguation for Autonomous Vehicles

Saumya Kumaar[1], Navaneethkrishnan B[1], Sumedh Mannar[1] and S N Omkar[1]

*Abstract*— With a great amount of research going on in the field of autonomous vehicles or self-driving cars, there has been considerable progress in road detection and tracking algorithms. Most of these algorithms use GPS to handle road junctions and its subsequent decisions. However, there are places in the urban environment where it becomes difficult to get GPS fixes which render the junction decision handling erroneous or possibly risky. Vision-based junction detection, however, does not have such problems. This paper proposes a novel deep convolutional neural network architecture for disambiguation of junctions from roads with a high degree of accuracy. This network is benchmarked against other well known classifying network architectures like AlexNet and VGGnet. Further, we discuss a potential road navigation methodology which uses the proposed network model. We conclude by performing an experimental validation of the trained network and the navigational method on the roads of the Indian Institute of Science (IISc).

## I. INTRODUCTION

Autonomy in transportation systems have garnered a great level of interest in both academic and industrial research communities in the past few years [1] Apart from the fully autonomous prototypes, the automobile industry has been equipping their products with intelligent assistive features like Lane tracking, LiDAR-based braking, and the parallel-parking assist. Despite exponential advances in the development of these intelligent features, the world is yet to witness a fully-autonomous system in service. The main hurdle faced in dealing with such autonomous systems is the development of its navigational system. Earliest developments in autonomous navigation of ground vehicles involved point to point navigation using Global Positioning System (GPS)[2]. However, the signal strength, being in the order of $10^{-17}$ Watts, becomes susceptible to interferences which degrades the quality of the signal especially under tree canopies, tunnels, and building basements. K.Furuno et.al. [3] proposed a navigational system that overcomes the limitations posed by the GPS-based methods. This method comprises of a number of guidance information transmitter systems dispersed along the route of interest, where each transmitter broadcasts the current location information in the form of extremely-low-powered electric waves, and a guidance signal generator system for providing guidance information for directing towards the next transmitter in the same route. A similar method was proposed by T.Saito [4]. This method, that uses radio-beacons mounted on street lights, shows promising results. Despite the glorious advantages such methods provide, the implementation of such systems proves to be a costly affair with recurrent maintenance costs apart from the initial installation cost.

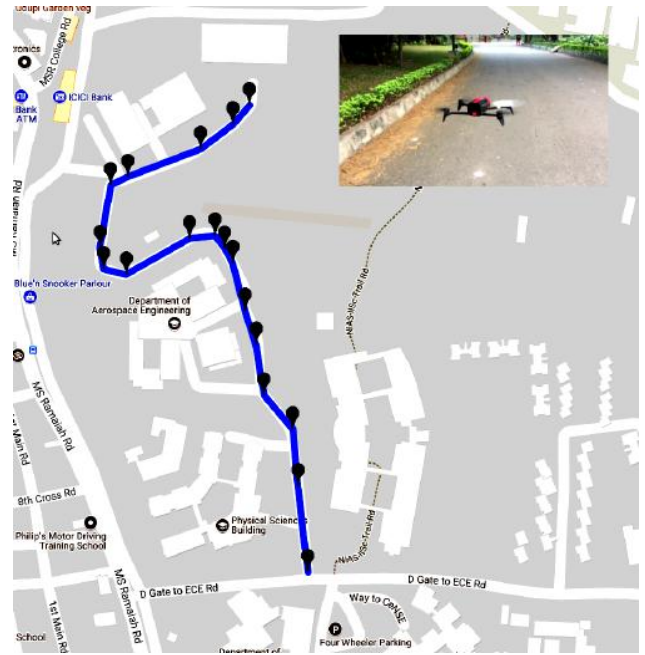

Fig. 1: Due to their ability to serve as a rapid-prototyping platform, off-the-shelf drones are commonly used to test autonomous vehicle navigation algorithms.This is GPS Flight Data of a test conducted at IISc,Bangalore, overlayed on Google Map.This achieved using JuncNet integrated with our in-house road following algorithm. It detects the junctions and takes only requisite turns so as to complete the desired track.

Due to their low cost, research on vision-based navigation methods has taken giant strides in the past decade. One of the first few pieces of research in this field involved texture based identification and extraction of road pixels using monocular vision [5,6,7]. Nowadays, with the advent of powerful computing technology, road segmentation is approached as a machine learning problem and is being seamlessly integrated into vehicles for road/lane tracking[8,9,10,11]. Both learning and non-learning methods rely on the driving control based on the detected road pixels and aligning the vehicle to the center of the detected road cluster. Though there are a lot of methods to navigate roads, there are no research papers that deal with road junctions in real time which are of absolute necessity in the context of navigation in an urban setting. D. Bhatt *et.al*[12] evaluated the use of Long-term Recurrent Convolutional Networks (LRCN) but it was not proven in real-time implementation. Another drawback the research had was the training was done on merely a total of 372 images which is not enough to evaluate its proficiency in classification. Most of the other existing research on road and junction classification has been performed on the aerial imagery used in Low-Altitude Remote Sensing (LARS) studies. Hence, there exists a need for a real-time road intersection detector in the context of autonomous vehicular navigation.

[1] All the authors are with the Indian Institute of Science, Bangalore
* This project was funded by the Robert Bosch Center for Cyber Physical Systems, Indian Institute of Science, Bangalore

The primary contributions of this paper are:

(1) We propose an experimental convolutional neural network (CNN) architecture and training procedure for a high accuracy binary classifier for road junction disambiguation

(2) We also discuss a navigation methodology where the proposed architecture can be used in conjunction with any existing road tracking algorithm and the real-time implementation and its integrability with other navigation algorithms is also demonstrated

The experimental verification of the proposed network architecture and the navigation methodology is implemented on an off-the-shelf unmanned aerial vehicle (UAV) call Bebop 2 manufactured by Parrot. Drones enable rapid prototyping of algorithms meant for autonomous cars as the equipment required to implement the same on vehicular platform tend to be expensive. UAVs being comparatively cheap, provide the right perspective to obtain the necessary input frames without incurring heavy contingency losses in case of untoward incidents.

## II. METHODOLOGY

The junction detection algorithm presented in this paper has a total of 5 stages as shown in the flow diagram (Fig.2). This section is further divided amongst the same.

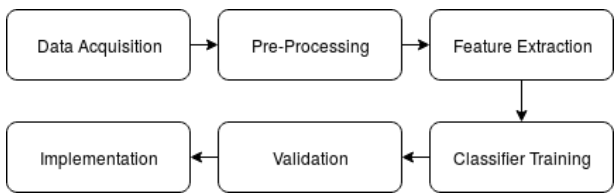

Fig. 2: Flow of the Algorithm

### A. Image Acquisition and Preprocessing

A Bebop Parrot 2 was used in the research for data acquisition as well as for implementation. Data is streamed in standard H.264 format (1920×1080 resolution) which is then resized later for further processing. Videos were taken of every junction for 30 seconds and the individual frames were extracted for processing.

Since the presented research deals with the problem of road-junction disambiguation, we apply an image sharpening filter on the dataset. The regular roads usually have edge energies, due to the presence of boundaries and the same are enhanced even more with image sharpening, as shown in Fig.3. The road junctions are usually widespread, so the boundary energies are not very prominent in junction images.

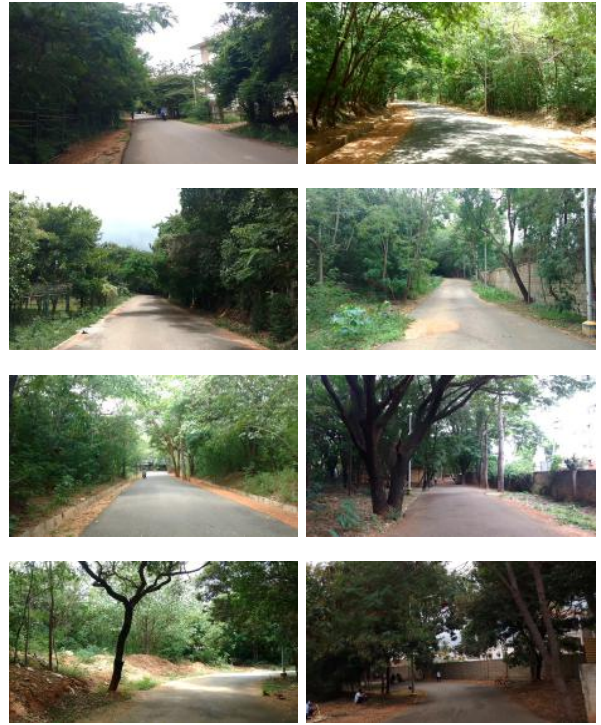

Fig. 3: Straight roads custom dataset collected in and around IISc Bangalore. The dataset helps in the classification of *none* from the *junctions*

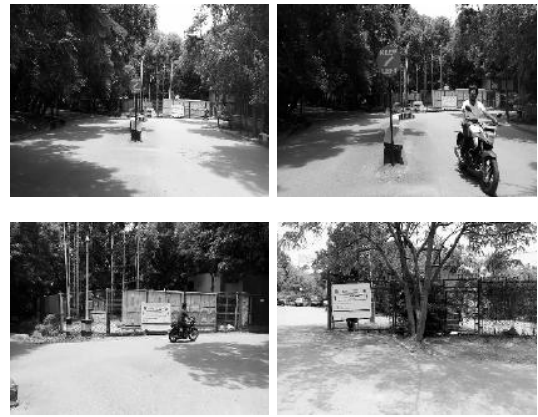

Fig. 4: Some of the grayscale road-junction images taken in and around IISc Bangalore campus.

Once the edge energies are enhanced, our JuncNet model learns them and predicts the presence or absence of a road junction.

The dataset comprises of two aspects: Custom (Fig. 3 and Fig.4) and ImageNet. The custom dataset has been collected in and around the Indian Institute of Science, Bengaluru campus whereas the ImageNet junction dataset is universal. We validate on both.

### B. Feature Extraction

We propose to use to Radon Features for road-junction disambiguation problem as, unlike the standard feature ex-traction techniques like Canny-Edge, Histogram of oriented gradients (HOG), Speeded-up robust features (SURF) etc., Radon Transformation computes features based on cumulative line-integrals of pixels that result in good road-segmentation despite feature-rich elements like mud or light shadows. Radon Features are primarily meant to segment edge energies while being tolerant to aberrations. The reconstructed images show the edge-filled regions as white patches whereas the black patches indicate the non-edge regions, as shown in the figures below. The Radon transform is a generic mathematical transformation technique commonly used for CAT scan analysis, but this research is focused on reconstructing the sinograms after the application of radon transformation to extract required features, the process of which is described below :

*1) Reconstruction Approach - Ill-posedness:* The process of reconstruction (here done using Ill-posedness) helps in the construction of the image (or the equivalent function $f$) from the projection information. Reconstruction is tackled as an inverse problem.

$$c_n f = (-\Delta)^{(n-1)/2} R^* R f \quad (1)$$

where $Rf$ is the radon transformed image matrix, $R^*$ is the adjoint of $Rf$ and ,

$$c_n = (4\pi)^{(n-1)/2} \frac{\Gamma(n/2)}{\Gamma(1/2)} \quad (2)$$

and the power of the Laplacian $-\Delta^{(n-1)/2}$ is defined as a pseudo-differential operator if necessary by the Fourier transform :

$$F\left[-\Delta^{(n-1)/2}\phi\right](\xi) = |2\pi\xi|^{n-1} F\phi(\xi) \quad (3)$$

For computational purposes and efficiency, the power of the Laplacian is commuted with the dual transform $R^*$ to give

$$c_n f = \begin{cases} R^* \frac{d^{n-1}}{ds^{n-1}} Rf, & n \text{ odd} \\ R^* H_s \frac{d^{n-1}}{ds^{n-1}} Rf, & n \text{ even} \end{cases}$$

where $H_s$ is the Hilbert transformation matrix with respect to the $s$ variable. In two dimensions, the operator $H_s d/ds$ commonly appears in image processing techniques and video processing techniques as a *ramp filter*. We can therefore, prove directly from the Fourier slice theorem and change of variables for integration that for a compactly supported continuous function of two variables the following holds true :

$$f = \frac{1}{2} R * H_s \frac{d}{ds} Rf \quad (4)$$

Thus in an image processing/video processing scenario the original image/frame can be reconstructed from the *sinogram* data $R$ by applying a fundamental ramp filter (in the $s$ variable) and then back-projecting as discussed previously. As the filtering step can be performed efficiently and effectively (for example using digital signal processing techniques) and the back projection step is simply an conglomeration of values in the pixels of the image, this results in a highly efficient, and hence widely used, algorithm.

Explicitly, the inversion formula obtained by the latter method is :

$$f(x) = \frac{1}{2}(2\pi)^{1-n}(-1)^{(n-1)/2} \int_{S^{n-1}} \frac{\partial^{n-1}}{\partial s^{n-1}} Rf(\alpha, \alpha \cdot x) d\alpha \quad (5)$$

if $n$ is odd, and

$$f(x) = \frac{1}{2}(2\pi)^{-n}(-1)^{n/2} \int_{-\infty}^{\infty} \frac{1}{q} \int_{S^{n-1}} \frac{\partial^{n-1}}{\partial s^{n-1}} Rf(\alpha, \alpha \cdot x + q) d\alpha dq \quad (6)$$

if $n$ is even.

So the the generic Radon transform-and-reconstruction when applied to our road-junction dataset is observed as follows. As is seen clearly, the road-junctions appear as extensive *black* regions in the images whereas the other side-paths appear as *white*. These grayscale images along with the radon reconstructed image (Fig. 5) are then fed to the neural network for classification.

*C. The JuncNet Model*

For the classification, a 2-CNN / 2-FCN (with one output layer) architecture(Fig.6) is used, with input image size of 64 X 64 and a batch size of 30 (for a real-time implementation). The learning rate of the network was fixed at $10^{-5}$, and the categorical cross-entropy loss was reduced using the standard Adam Optimizer Function, running for 200 epochs. The AlexNet architecture was also tested on both custom and Image-net dataset, but the prediction metrics are evaluated for CIFAR-10/100 datasets, rather than any generic dataset. The training of the network was carried out on a system with specifications listed in Table II.

TABLE I: System Specifications

| Hardware | Specification |
| --- | --- |
| Memory | 32 GB |
| Processor | Intel Core i7-4770 CPU @ 3.4 GHz x 8 |
| GPU | GeForce GTX 750 Ti/PCIe/SSE2 |
| OS Type | 64-bit Ubuntu 16.04 LTS |

III. HARDWARE IMPLEMENTATION

An off-the-shelf quadrotor, Parrot Bebop 2 (Fig. 7), with open-source control stations available, was used as hardware platform for algorithm development and testing. There is no on-board computation taking place on the drone's hardware, everything occurs in the user's laptop and only high level velocity and yaw commands are published to it via WiFi.

*A. Navigation Algorithm*

Once the frame is classified to have a road-intersection, the algorithm keeps a count of how many junctions have been identified. The UAV is pre-programmed to take turns at appropriate junctions corresponding to the count. For example, if $3^{rd}$ junction is identified, the drone makes an appropriate yaw movement to align itself with the next patch of road to be followed. So this model is designed to be modular and works as an add-on to other road-navigation systems.

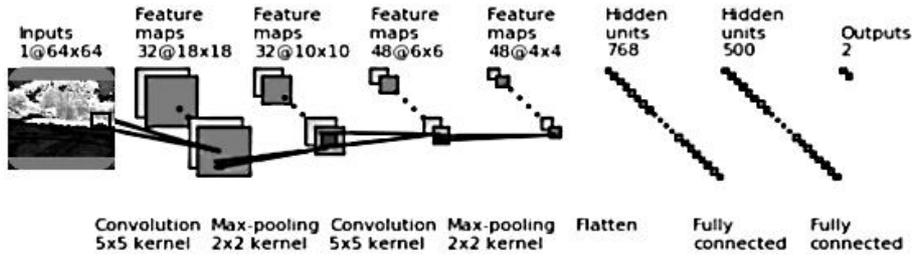

Fig. 6: The JuncNet consists of two convolutional layers and 2 Dense layers sandwiched by two MaxPooling layers.

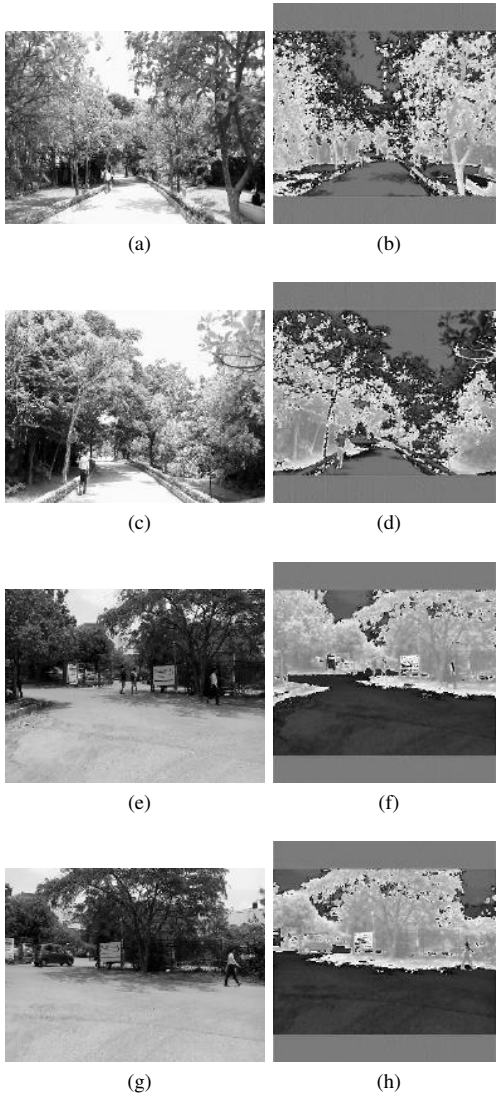

Fig. 5: Straight roads and junctions in the frames on the left are clearly distinguished in the Radon reconstructed images as seen on the right. The above images belong to the custom dataset

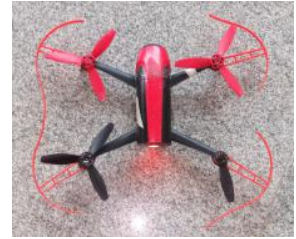

Fig. 7: Parrot Bebop 2

## IV. EVALUATION AND RESULTS

Since there are not many metrics available pertaining to our current problem statement, we report the classification accuracy of our JuncNet on custom dataset as well as dataset from ImageNet. We also verify against some of the very famous classification networks like the AlexNet and VGGNet for further validation on the same dataset. The algorithm has been experimentally tested to work at 25 FPS, which is suitable for real-time implementation.

Moreover, the junction disambiguation problem presented in this research is in the form of a binary classification. The system tries to predict whether a junction is present in the frame or not. The program has calculated number of junctions and every junction is associated with a suitable yaw command that is sent to the drone to take the appropriate turn. So, the neural network is tuned for image classification which is why it has been tested on MNIST and CIFAR-10 datasets, which are the standard benchmarks for classification.

As stated previously, JuncNet has been tested as an add-on feature to other road-tracking and following algorithms. Fig. 1 shows the GPS plot of the road strip that our UAV followed while taking decisions about the junction. Since the desired path was straight, the drone was programmed appropriately to avoid turning at all junctions.(Fig.7)

### A. MNIST

The database of handwritten digits provided by the Modified National Institute of Standards and Technology (MNIST), provides a training set of 60,000 examples, and a test set of 10,000 images. The digit images have been size-normalized and and have been put in the center in a fixed-size image of size 28×28 pixels. It is a standard database for benchmarking learning techniques, pattern analysis and recognition methods.

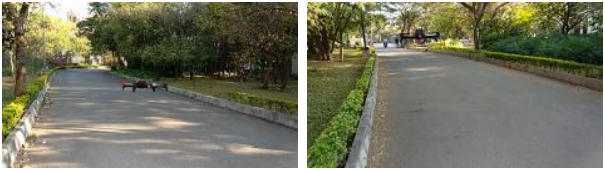

Fig. 8: JuncNet being used in conjunction with a road-following and tracking algorithm. JuncNet allows junction disambiguation even in the absence of GPS.

TABLE II: JuncNet Classification Results on MNIST

| Architecture | Error Rate |
|---|---|
| APAC [13] | 0.23% |
| C-SVDDNet [14] | 0.35% |
| ReNet [15] | 0.45% |
| **JuncNet** | 0.55% |
| Invariant SVM [17] | 0.56% |
| PCANet [16] | 0.62% |

We have compared our algorithm against many of the established classification techniques. The classification outcomes are tabulated in Table II.

### B. CIFAR-10

The CIFAR-10 database of images (Canadian Institute For Advanced Research) is a conglomeration of images that are commonly used to train machine learning and computer vision algorithms. It is one of the most commonly used datasets for deep learning and machine learning research. The CIFAR-10 dataset comprises of 60,000 32×32 color images in 10 different categories. There are 6000 images of each category and there are 50000 training images and 10000 test images. This database is used to benchmark the networks' performance in multi-class classification problems. The classification outcomes have been tabulated in Table III.

TABLE III: JuncNet Classification Results on CIFAR-10

| Architecture | Accuracy |
|---|---|
| APAC [13] | 89.67% |
| ReNet [15] | 87.65% |
| DeepNet [18] | 89% |
| **JuncNet** | 83.15% |
| Sinlge-Layer Net [19] | 81.56% |
| PCANet [16] | 78.67% |

### C. Custom Dataset

In order to test the robustness of our algorithm, we have tested it in various configurations, like predicting the presence/absence of a junction in an image and predicting the class of the junction itself. The results (Fig. 8) make the stance more clear. All roads are marked as *none* and rest is classified as *junction*.

Furthermore, this algorithm can also be extended to classification of different junctions. As of now, it only predicts the presence of junctions in an image, but results presented in Fig. 9 shows that **JuncNet** is capable of junction discrimination, provided that road-intersection images are taken from different angles. This can be used to predict the location of the drone based on the label of the junction that is identified.

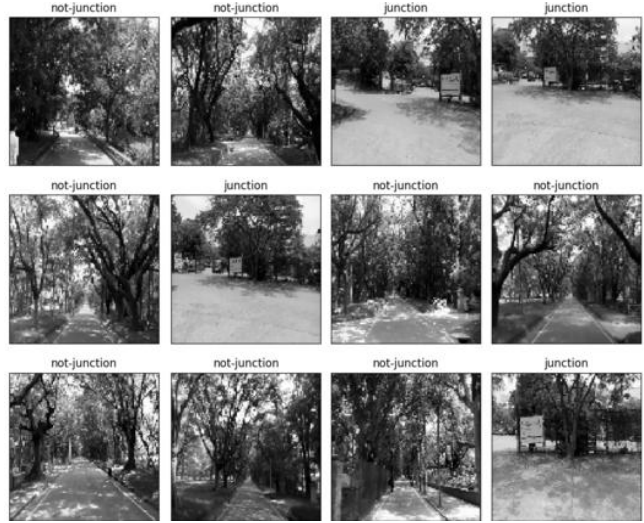

Fig. 9: Examples of Junction/None classification on custom dataset

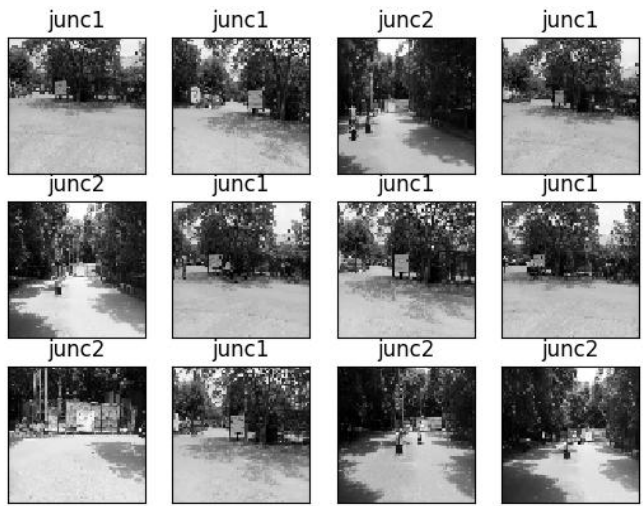

Fig. 10: Junction1-Junction2 Classification by JuncNet

TABLE IV: Algorithms and Accuracies

| Algorithm | Image-Net | Custom Dataset |
|---|---|---|
| kNN | 52.45 % | 48.87 % |
| SVM | 74.34 % | 68.87 % |
| 1-CNN/2-FCN | 90.12 % | 91.22 % |
| **JuncNet** | **98.74** % | **97.34** % |
| AlexNet | 94.22 % | 94.34 % |

### D. Image-Net Dataset

The robustness of our algorithm was established by cross-validating it on the image-net road-junction dataset as seen

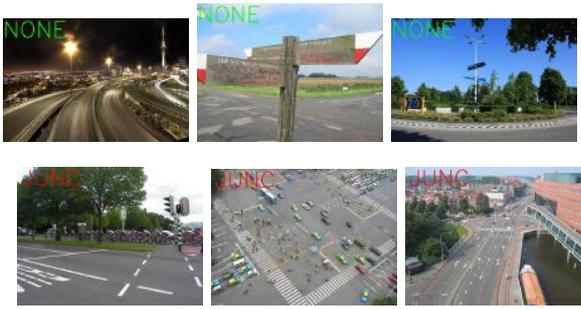

Fig. 11: Junction/None Classification on the dataset from Image-Net. Misclassifications were noted in this dataset due to the presence of obstacles in front of road junctions and also due to insufficient number of images available to train from this particular dataset

in Fig. 10.

## V. Conclusion

An alternate approach to junction-disambiguation has been discussed and presented in the research. Although most of the road-junction identification techniques worked from the BEV perspective, this approach takes the camera in-line-of-view one. The future work of this research involves improving the architecture for even more fine-tuned results of junction disambiguation. Required dataset collection is also a task that needs to be addressed. This approach, although experimentally verified on Unmanned Aerial Vehicles with VTOL capabilities, could also be extended to ground rovers and self-driving cars.

## References

[1] Sebastian Thrun. Toward robotic cars. *Commun. ACM*, 53(4):99106, April 2010.
[2] Abbott, Eric, and David Powell. "Land-vehicle navigation using GPS." *Proceedings of the IEEE 87*, no. 1 (1999): 145-162.
[3] Furuno, Kenichi, Yoshio Tooko, and Takayuki Nonami. "Road navigation system." *U.S. Patent 4,819,174*, issued April 4, 1989.
[4] Saito, Takaharu, Junkoh Shima, H. Kanemitsu, and Y. Tanaka. "Automobile navigation system using beacon information." *In Vehicle Navigation and Information Systems Conference, 1989. Conference Record*, pp. 139-145. IEEE, 1989.
[5] Kuan, Darwin, Gary Phipps, and A-C. Hsueh. "Autonomous robotic vehicle road following." *IEEE Transactions on Pattern Analysis and Machine Intelligence 10*, no. 5 (1988): 648-658.
[6] He, Yinghua, Hong Wang, and Bo Zhang. "Color-based road detection in urban traffic scenes." *IEEE Transactions on intelligent transportation systems 5*, no. 4 (2004): 309-318.
[7] Broggi, Alberto, and Simona Berte. "Vision-based road detection in automotive systems: A real-time expectation-driven approach." *Journal of Artificial Intelligence Research 3* (1995): 325-348.
[8] Zhou, Shengyan, Jianwei Gong, Guangming Xiong, Huiyan Chen, and Karl Iagnemma. "Road detection using support vector machine based on online learning and evaluation." In *Intelligent Vehicles Symposium (IV), 2010 IEEE*, pp. 256-261. IEEE, 2010.
[9] Guo, Chunzhao, Seiichi Mita, and David McAllester. "Robust road detection and tracking in challenging scenarios based on Markov random fields with unsupervised learning." *IEEE Transactions on intelligent transportation systems* 13, no. 3 (2012): 1338-1354.
[10] Alvarez, Jos M. lvarez, and Antonio M. Lopez. "Road detection based on illuminant invariance." *IEEE Transactions on Intelligent Transportation Systems* 12, no. 1 (2011): 184-193.
[11] Foedisch, Mike, and Aya Takeuchi. "Adaptive real-time road detection using neural networks." *In Intelligent Transportation Systems, 2004. Proceedings. The 7th International IEEE Conference on*, pp. 167-172. IEEE, 2004.
[12] Bhatt, Dhaivat, Danish Sodhi, Arghya Pal, Vineeth Balasubramanian, and Madhava Krishna. "Have I Reached the Intersection: A Deep Learning-Based Approach for Intersection Detection from Monocular Cameras." (2017).
[13] Sato, Ikuro, Hiroki Nishimura, and Kensuke Yokoi. "Apac: Augmented pattern classification with neural networks." *arXiv* preprint arXiv:1505.03229 (2015).
[14] Wang, Dong, and Xiaoyang Tan. "Unsupervised feature learning with c-svddnet." Pattern Recognition 60 (2016): 473-485.
[15] Visin, Francesco, Kyle Kastner, Kyunghyun Cho, Matteo Matteucci, Aaron Courville, and Yoshua Bengio. "Renet: A recurrent neural network based alternative to convolutional networks." *arXiv* preprint arXiv:1505.00393 (2015).
[16] Chan, Tsung-Han, Kui Jia, Shenghua Gao, Jiwen Lu, Zinan Zeng, and Yi Ma. "PCANet: A simple deep learning baseline for image classification?." *IEEE Transactions on Image Processing 24*, no. 12 (2015): 5017-5032.
[17] Decoste, Dennis, and Bernhard Schlkopf. "Training invariant support vector machines." *Machine learning 4*6, no. 1-3 (2002): 161-190.
[18] Krizhevsky, Alex, Ilya Sutskever, and Geoffrey E. Hinton. "Imagenet classification with deep convolutional neural networks." *In Advances in neural information processing systems*, pp. 1097-1105. 2012.
[19] Coates, Adam, Andrew Ng, and Honglak Lee. "An analysis of single-layer networks in unsupervised feature learning." *In Proceedings of the fourteenth international conference on artificial intelligence and statistics*, pp. 215-223. 2011.